\documentclass[letterpaper, 10 pt, conference]{ieeeconf}  

\usepackage{graphicx}
\usepackage{color}
\usepackage{amsfonts}
\usepackage{cite}
\usepackage[font={footnotesize,it}]{caption}
\usepackage[dvipsnames]{xcolor}
\usepackage{subcaption}
\usepackage{graphicx,subcaption}
\usepackage{amsmath} 
\usepackage{amssymb}  
\usepackage{tabularx}
\usepackage{multirow}

\usepackage[textsize=tiny]{todonotes}

\makeatletter
\newcommand{\thickhline}{%
    \noalign {\ifnum 0=`}\fi \hrule height 1pt
    \futurelet \reserved@a \@xhline
}
\newcolumntype{"}{@{\hskip\tabcolsep\vrule width 1pt\hskip\tabcolsep}}
\makeatother

\IEEEoverridecommandlockouts                              

\title{\LARGE \bf
Offline Goal-Conditioned Reinforcement Learning for Shape Control of Deformable Linear Objects
}

\author{Rita Laezza$^{1}$, Mohammadreza Shetab-Bushehri$^{2}$, Gabriel Arslan Waltersson$^{1}$, \\ Erol \"{O}zg\"{u}r$^{2}$, Youcef Mezouar$^{2}$ and Yiannis Karayiannidis$^{3}$
\thanks{$^{1}$ Division of Systems and Control, Department of Electrical Engineering, Chalmers University of Technology, Sweden
        ({\tt \{laezza, gabwal\}@chalmers.se})}%
\thanks{$^{2}$ CNRS, Clermont Auvergne INP, Institut Pascal, Universit\'{e} Clermont Auvergne, France ({\tt\{m.r.shetab, }
        {\tt erolozgur\}@gmail.com, youcef.mezouar@sigma-clermont.fr)}}%
\thanks{$^{3}$ Department of Automatic Control, Lund University, Sweden ({\tt\small  yiannis@control.lth.se}.) The author is a member of the ELLIIT Strategic Research Area at Lund University.}%
}
\begin{document}

\maketitle
\thispagestyle{empty}
\pagestyle{empty}

\begin{abstract}
Deformable objects present several challenges to the field of robotic manipulation. One of the tasks that best encapsulates the difficulties arising due to non-rigid behavior is \textit{shape control}, which requires driving an object to a desired shape. 
While shape-servoing methods have been shown successful in contexts with approximately linear behavior, they can fail in tasks with more complex dynamics. We investigate an alternative approach, using offline RL to solve a planar shape control problem of a Deformable Linear Object (DLO). To evaluate the effect of material properties, two DLOs are tested namely a soft rope and an elastic cord. We frame this task as a goal-conditioned offline RL problem, and aim to learn to generalize to unseen goal shapes. Data collection and augmentation procedures are proposed to limit the amount of experimental data which needs to be collected with the real robot. We evaluate the amount of augmentation needed to achieve the best results, and test the effect of regularization through behavior cloning on the TD3+BC algorithm. Finally, we show that the proposed approach is able to outperform a shape-servoing baseline in a curvature inversion experiment. 
\end{abstract}


\section{INTRODUCTION}
\label{sec:introduction} 
As collaborative robots become widespread, more industries begin to explore automation, bringing to light several manipulation skills which are yet to mature in the field of robotics. For instance, new problems from agriculture, food processing and healthcare, beyond many problems found in manufacturing, can involve skillful manipulation of deformable objects \cite{zhu2022challenges}.
Open challenges in this field stem from a few notable characteristics of deformable objects, namely \textbf{(i)} high degrees of freedom which are underactuated and difficult to both sense and control \textbf{(ii)} diverse material properties that lead to a wide range of deformation behaviors, and \textbf{(iii)} the high occurrence of visual self-occlusions which cause ambiguities during tracking.  

Shape control is a type of manipulation problem which is unique to deformable objects, where the goal is not only to change the position of an object, but also its shape. In contrast, when manipulating rigid objects, the objective can usually be reduced to a desired pose. The classical approach to solve this problem is referred to as shape-servoing, which is characterized by tasks where the goal is to move controlled points on the object to achieve an explicit shape \cite{berenson2013manipulation, aranda2020monocular, shetab2022rigid, Shetab2022lattice}. Despite significant progress, classical methods suffer from several limitations, including computational complexity and modeling inaccuracy, e.g. due to difficulties in identifying the mechanical parameters of the object and its interaction with the environment. Furthermore, such methods mostly rely on an instantaneous error and local models, therefore objects with complex material properties remain to be explored, since their manipulation exhibits more long-term effects. 

\begin{figure}[!t]\centering
    \vspace{0.3cm}
    \includegraphics[width=\linewidth]{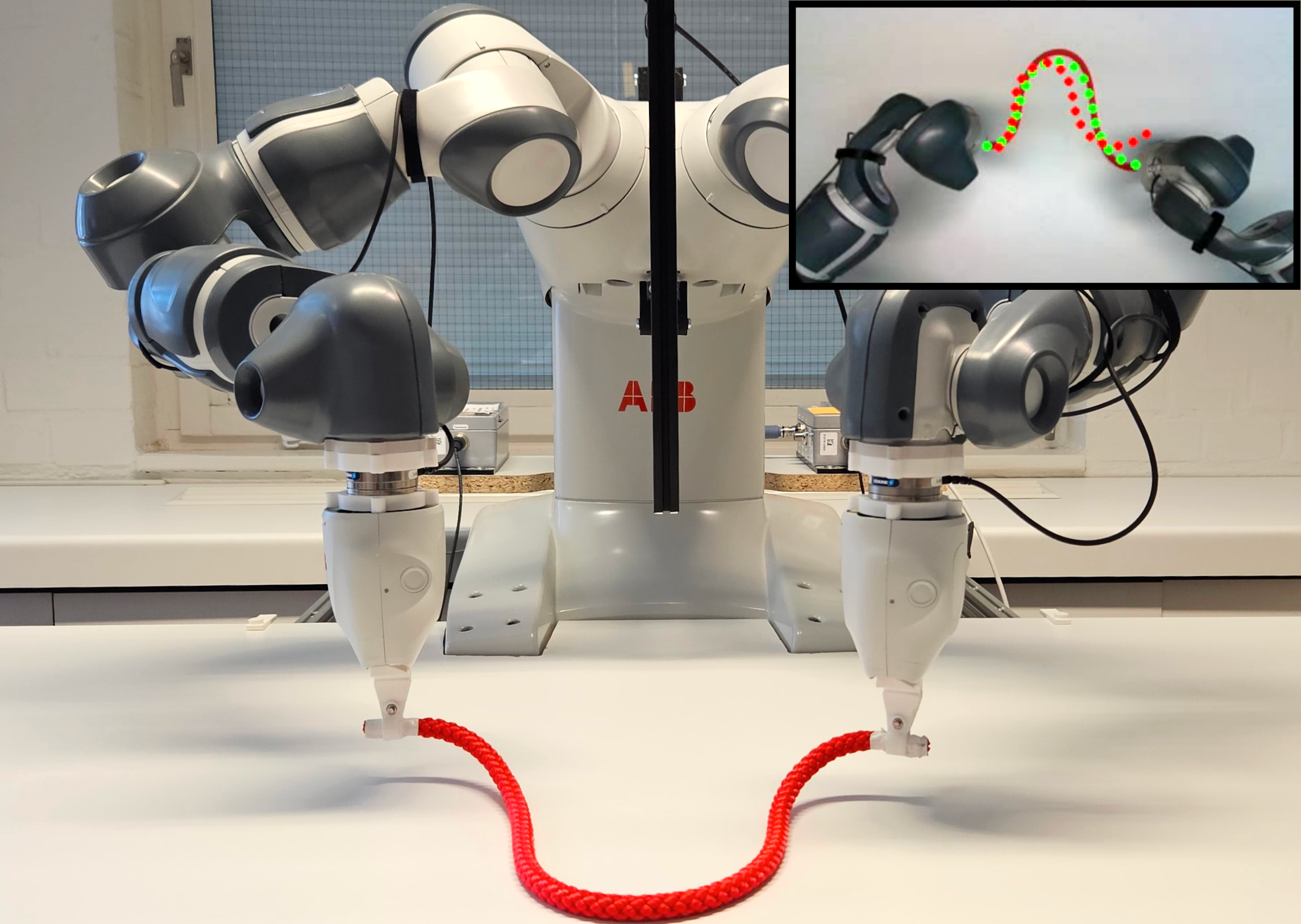} 
    \caption{Dual-arm ABB YuMi robot manipulating a DLO on a table. A fixed Intel RealSense camera provides a top-view of the workspace, with the field of view shown in the top right corner. The DLO state is shown a set of points, with the current shape tracking in green and the goal shape in red.}
    \label{fig:setup} 
    \vspace{-0.2cm}
\end{figure}

\begin{figure*}[!t]
    \centering
    \vspace{0.2cm}
    \includegraphics[width=0.98\linewidth]{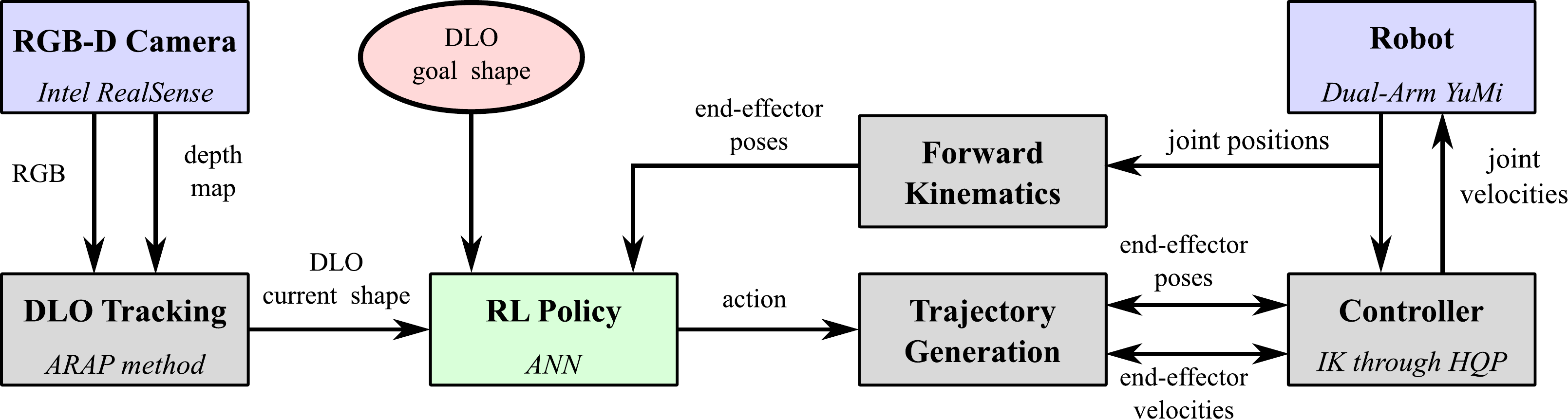} 
    \caption{Block diagram showing system architecture. Blue blocks indicate hardware components, i.e. the camera and the robot. The green block is the \textbf{RL Policy} modeled as an ANN, which is trained using the proposed method. Red indicates the user input, namely a desired shape of the DLO. Finally, gray blocks constitute the other software components for perception i.e. \textbf{DLO Tracking}, and control, i.e. \textbf{Trajectory Generation}, \textbf{Forward Kinematics} and \textbf{Controller}. A PD velocity controller is used to track the generated end-effector velocity trajectories, based on a given RL policy action. If a new action is received before the previous is completed, the new trajectories use the current end-effector poses and velocities as a starting point. Joint velocities are computed via an inverse kinematics formulation, solved through HQP optimization.}
    \label{fig:schematics}
    \vspace{-0.2cm}
\end{figure*}
Recent progress in deformable object manipulation has been fueled by advances in Deep Learning (DL) research \cite{matas2018sim, jangir2020dynamic,magassouba2023multi}. DL methods have the advantage of indirectly encoding the dynamics of the manipulation task, without requiring extensive engineering work in order to model the object-environment interaction. This is appealing due to the large variety of deformation behaviors across different classes of deformable objects \cite{sanchez2018robotic}. Reinforcement Learning (RL) in particular, allows a robot to learn from experience and optimize a policy towards long-term objectives. However, a key obstacle to applying RL in real-world applications is the need to collect online data on a robotic setup, which is time consuming and potentially unsafe due the stochastic nature of RL exploration. An alternative is to first collect real data using some heuristic policy and then use it to train a policy through offline RL algorithms \cite{levine2020offline}. 

While RL is usually formulated with respect to a unique goal, there are many robotic tasks that are best described as multi-goal. For example, in a request for more research, Plappert et al. \cite{plappert2018multi} present several robotic simulation environments based on a 7-DoF Fetch  manipulator with simple multi-goal tasks, such as pick-and-place. Even though this task is easily solvable with model-based control, it is still challenging for RL-based approaches. Indeed when presenting the Hindsight Experience Replay (HER) algorithm for multi-goal problems, Andrychowicz et al. \cite{andrychowicz2017hindsight} show that without their contribution, plain RL algorithms cannot solve the task. This form of RL is also referred to a Goal-Conditioned Reinforcement Learning (GCRL), since it usually involves rewards and policies which are conditioned on the goal that is to be achieved in a given episode \cite{ijcai2022p770}. 

In this work we tackle a Deformable Linear Object (DLO) shape control problem with surface interactions, shown in Fig. \ref{fig:setup}. To validate our modular approach we perform experiments on both a soft rope and an elastic cord, which have significantly different material properties. We investigate the efficacy of offline RL, using the TD3+BC algorithm proposed by Fujimoto et al. \cite{fujimoto2021minimalist}. We further compare the learned control policies with a well-established shape-servoing method proposed by Berenson \cite{berenson2013manipulation}. Additionally, we show that a simple data augmentation approach inspired by the HER principle improves performance. To the best of our knowledge this is the first implementation of offline GCRL on a DLO shape control problem, entirely executed using real data and robot experiments.

\section{RELATED WORK}
\label{sec:related_work} 
Recent work by Almaghout et al. \cite{almaghout2024robotic} offers a comprehensive review of the main approaches found in the literature to tackle DLO shape control problems. They classify the DLO manipulation strategies into four types: \textbf{(i)} model-based, \textbf{(ii)} Jacobian-based, \textbf{(iii)} data-driven and \textbf{(iv)} hybrid. Note that the first two approaches can both be considered as model-based, with the key difference being that in \textbf{(i)} a physical model is used explicitly to simulate the dynamics of the DLO, while in \textbf{(ii)} the Jacobian provides an approximate model of the motion of the DLO with respect to the movement of the end-effectors. Jacobian-based approaches are the most commonly used to solve shape-servoing problems, with two main strategies for obtaining the Jacobian of the DLO: numerical or analytical. Works by Zhu et al. \cite{zhu2018dual}, Jin et al. \cite{jin2019robust} and Lagneau et al. \cite{lagneau2020automatic} are all examples of numerical strategies. On the other hand, works by Berenson \cite{berenson2013manipulation}, Shetab-Busheri et al.\cite{Shetab2022lattice} and Almaghout et al. \cite{almaghout2024robotic} are examples of analytical strategies. In this work we apply a data-driven method and compare it to a Jacobian-based approach.

There have been a few deformable object manipulation problems to which online RL was applied. Many have resorted to a combination of imitation learning and RL, such as Balaguer et al. \cite{balaguer2011combining} who tackled a momentum fold of a planar deformable object and Matas et al. \cite{matas2018sim}, who tackled three end-to-end cloth-manipulation tasks using Deep Deterministic Policy Gradient (DDPG), with some pre-training using Learning from Demonstrations (LfD). Conversely Wu et al. \cite{wu2019learning} solved pick-and-place tasks of deformable objects without any pre-training. Both of the latter two works made use of simulated data to train the policies and needed to address the sim-to-real gap. In contrast, we attempt to avoid such a gap by learning directly from real data and using offline RL. Han et al. \cite{han2017model} used model-based RL to solve a similar task to the one presented in this paper, where the DLO was to be manipulated with a single arm so that the two ends would meet. While all of these works tackled single-goal tasks, here we attempt to learn a multi-goal task. 

Since the first results in this paper were presented \cite{laezza2023offline}, there has been some work on GCRL for non-rigid material manipulation that is important to highlight here. Niu et al. \cite{niu2023goats} presented GOATS - Goal Sampling Adaptation for Scooping, an algorithm for goal-sampling based on the curriculum learning idea, i.e. starting with simpler goals and progressively increasing to more complex ones. They show improved results in simulation and validate them on a real-world experiment of water scooping. When testing policies learned in simulation directly on a real-robot, they report some sim-to-real gap for scooping goals with larger volumes of water. Sun et al. \cite{dexdlo} on the other hand, tackled a DLO manipulation problem purely in simulation, using an end-to-end approach with a dexterous 24 Degrees of Freedom (DOFs) hand. Instead of learning an end-to-end policy, we present a modular system architecture (see Fig. \ref{fig:schematics}), where both perception and low-level control are removed from the learning problem. Furthermore, while both of these approaches made use of online GCRL in simulation environments, we apply offline GCRL purely on real data.

\section{PROBLEM STATEMENT}
\label{sec:problem_statement} 
\begin{figure}[!t]
    \vspace{0.2cm}
    \centering
    \includegraphics[width=\linewidth]{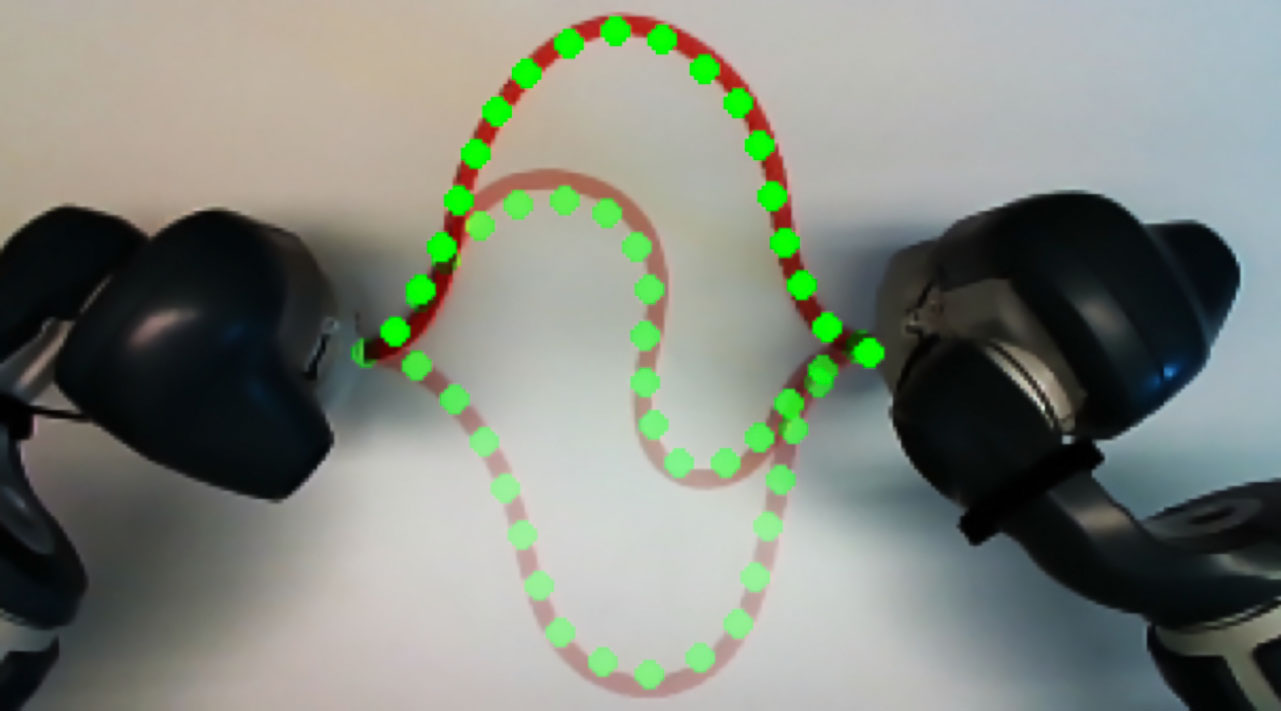} 
    \caption{Overlay of three possible DLO shapes with identical gripper poses. One can intuitively picture that for the top and bottom shapes, an additional top-down translation motion was necessary, while for the middle shape a counterclockwise rotation of the grippers must have occurred.}
    \label{fig:ambiguity}
    \vspace{-0.2cm}
\end{figure}

Prior work has tackled the problem of DLO shape control on a plane by two main approaches: \textbf{(i)} as a sequence of pick and place motions along the DLO with loose ends \cite{sundaresan2020learning, yan2020self, huo2021keypoint, caporali2024deformable}; or \textbf{(ii)} as continuous control problem with permanent grip of the DLO end-points by one or two end-effectors \cite{lagneau2020automatic, laezza2021learning, yang2022learning, yu2022global}. Caporali et al. \cite{caporali2024deformable} further differentiate the types of DLOs being manipulated as mostly \textit{soft} for approach \textbf{(i)} and \textit{elastic} for approach \textbf{(ii)}. This is an important observation, as indeed it becomes quite difficult to manipulate the shape of an elastic DLO (i.e. with high strain \cite{sanchez2018robotic}) with loose ends on a table, since its stiffness makes the object move almost rigidly. Conversely, moving a soft object (i.e. with low compression strength \cite{sanchez2018robotic}) with both ends constrained can lead to complex long-term effects which make this form of manipulation more challenging. In this work both types of DLOs are evaluated using approach \textbf{(ii)}.
 
 More specifically, the problem tackled in this work is the control of the task-space motion of two end-effectors holding the extremities of a DLO each with 3 DOFs, for translation on an $xy$-plane and orientation about the $z$-axis. Because the DLO is in contact with a surface, its dynamics are affected by friction. This problem formulation brings two main challenges: 
 {\def\theenumi{\alph{enumi}}
 \begin{enumerate}
     \item the middle part of the DLO is not directly affected by the movement of the end-effectors.
     \item target shapes may require moving away from the target to invert the curvature of the DLO, e.g. from concave to convex.
 \end{enumerate}}
 Notably, Almaghout et al. \cite{almaghout2024robotic} tackle the problem of curvature inversion of an elastic DLO, with relatively high stiffness, by proposing an Intermediary Shapes Generation (ISG) algorithm in combination with a Jacobian-based approach. A key property of this problem is that multiple DLO shapes can be achieved with the grippers in the same pose, depending on the preceding motion, as shown in Fig. \ref{fig:ambiguity}. This effect becomes more significant as the stiffness of the object decreases, i.e. from more elastic to softer DLOs.

\section{METHODS}
\label{sec:method} 
Robotic manipulation using RL methods often is formulated in an end-to-end manner, where a control policy is learned from raw image data to robot actuation signals, e.g. \cite{matas2018sim}. However, this increases the number of problems that have to be solved by the RL policy. In the particular case of shape servoing, the task of tracking the deformable object is challenging in itself and still an open research problem, e.g. \cite{tang2022track}. Therefore, in this work we assume the policy receives as input a simplified representation of the object, which is obtained using the state-of-the-art tracking algorithm proposed by Shetab-Bushehri et al. \cite{Shetab2022lattice}, based on an As-Rigid-As-Possible (ARAP) deformation model. Further, controlling the robot in joint space for this type of application is often unnecessary. Thus, a velocity controller is designed in task space and the Inverse Kinematics (IK) problem is solved through Hierarchical Quadratic Programming (HQP) to be able to add a hierarchy of constraints that keep the robot in a safe configuration \cite{escande2014hierarchical}. Fig. \ref{fig:schematics} shows a diagram of the proposed system architecture. The \textbf{DLO Tracking} algorithm is described in Section \ref{subsec:DLO_Tracking}, while Section \ref{subsec:controller} shortly describes the \textbf{Controller} design. Finally, Section \ref{subsec:rl} presents the proposed method to train the \textbf{RL Policy}.

\begin{figure}[t]
    \vspace{0.2cm}
    \centering
    \includegraphics[width=\linewidth]{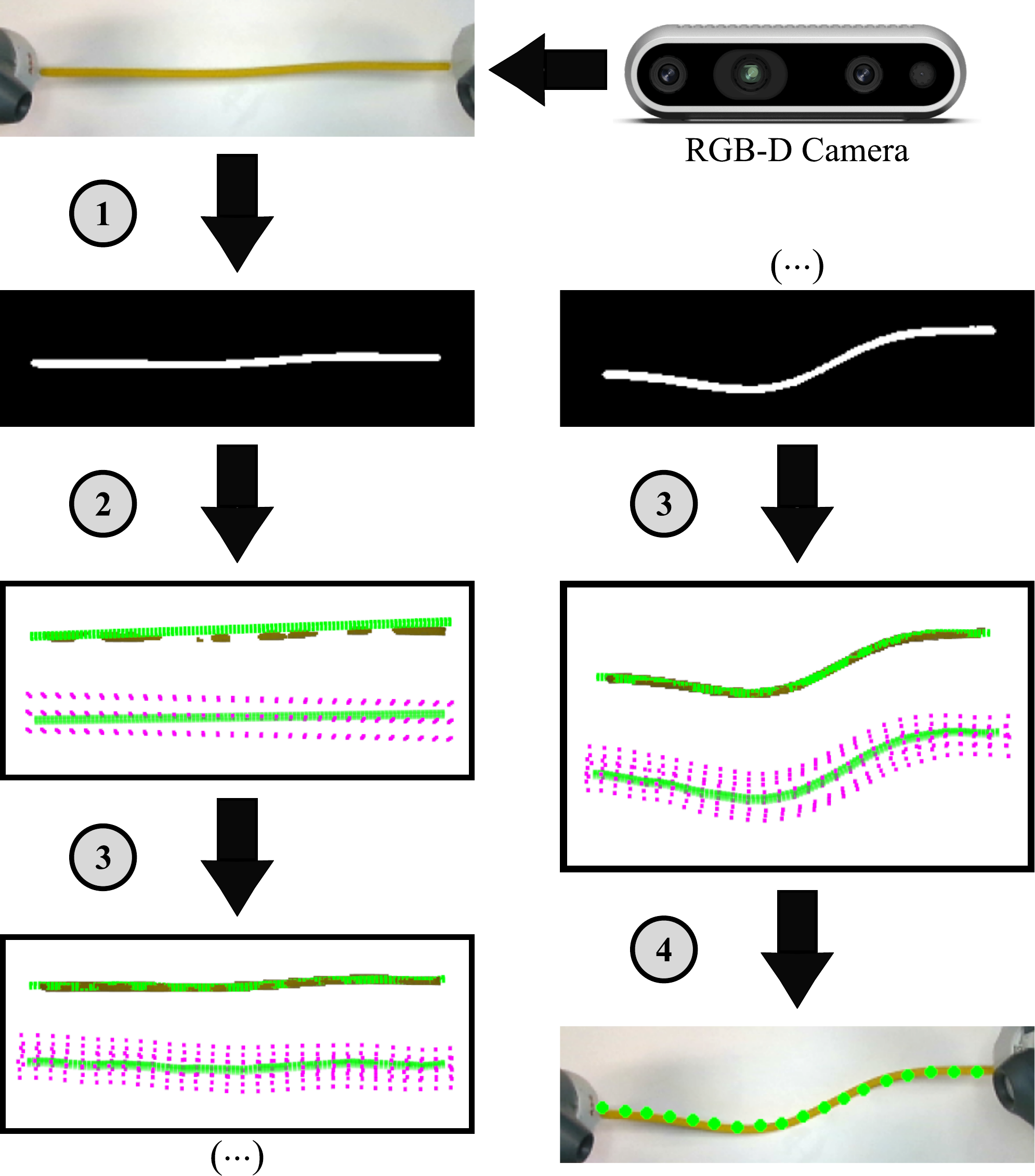} 
    \caption{Flowchart of DLO tracking procedure. A RGB-D camera provides the RGB feed shown at the top, but also the point cloud of the scene. The tracking procedure has the following sequence: \textbf{(1)} The RGB image is segmented based on the HSV value of the DLO; \textbf{(2)} the reference point cloud (green) is aligned with the point cloud of the DLO (yellow), and the surrounding lattice (pink) is constrained to the reference point cloud; \textbf{(3)} tracking begins by iteratively finding correspondences between the point cloud of the DLO and the reference, which in turn deforms the lattice using the ARAP principle; finally, \textbf{(4)} the points in the lattice are used to compute a set of N points along the DLO.}
    \label{fig:dlo_tracking}
    \vspace{-0.2cm}
\end{figure}

\subsection{DLO Tracking} 
\label{subsec:DLO_Tracking}
To track the DLO, we employ the algorithm developed by Shetab-Bushehri et al. \cite{Shetab2022lattice} illustrated in Fig. \ref{fig:dlo_tracking}. First, the RGB image is segmented to separate the DLO from the background using HSV filtering based on the hue value of the object (see image below item 1 in Fig. \ref{fig:dlo_tracking}). This segmentation is applied to obtain the point cloud of the DLO, by first aligning the depth information to the segmented image. Due to the small diameter of DLOs it can be challenging to separate the point cloud of the object from that of the table, often leading to gaps (see image below item 2 in Fig. \ref{fig:dlo_tracking}). 

Next, a reference point cloud, modeled as a semi-cylinder with the dimensions of the object, is aligned with the DLO's point cloud in its rest shape (i.e. a straight line). The tracking algorithm is initialized by forming a $3\times3\times M$ lattice around the reference point cloud of the DLO and then binding the two together by geometric constraints. During tracking, correspondences between the reference point cloud and the point cloud of the DLO are found in each frame. These corresponding points are applied as constraints to the lattice, effectively deforming it based on the ARAP deformation model \cite{alexa2000rigid}. As a result, the lattice's shape is updated and the DLO's reference point cloud becomes aligned with the captured point cloud, while keeping its local rigidity (see image below item 3 in Fig. \ref{fig:dlo_tracking}). 

Notably, the known state of the end-effectors is used to improve tracking, by constraining the positions of the lattice nodes closest to each end-effector based on their current poses. For this to work well, the relative camera pose must be obtained through external calibration. The final step to obtain the state of the DLO is to compute the mean coordinates of the lattice nodes belonging to each cross-section along its length, and then through an interpolation procedure obtain $N$ point coordinates evenly spaced along the DLO, with $N < M$ (see image below item 4 in Fig. \ref{fig:dlo_tracking}).


\subsection{Controller}
\label{subsec:controller}
As detailed in Section \ref{sec:problem_statement}, the DLO shape control problem is formulated in task space which requires an inverse kinematics controller to determine the desired joint velocity commands to send to the robot. First, a trajectory is generated for each arm by cubic interpolation based on desired end-effector poses and time constants, which is then used to obtain the \textit{desired motion} as end-effector velocities. Escande et al. \cite{escande2014hierarchical} proposed the HQP algorithm to solve IK problems as a hierarchy of equality and inequality constraints, faster than classical methods. We employ this method to track the desired end-effector velocity trajectories for each robot arm independently.  The HQP solver additionally takes feasibility objectives, based on the following hierarchy of constraints:
\begin{enumerate}
    \item joint velocity limit (inequality)
    \item joint position limit (inequality)
    \item elbow proximity limit (inequality)
    \item \textit{desired motion} (equality)
    \item joint position potential (equality)
\end{enumerate}
 The constraints with top priority keep the robot from exceeding joint velocity or position limits. Further, to avoid self-collisions the elbows are kept away from the robot body. After these safety constraints are satisfied, the desired motion in encoded as equality constraints using the Jacobian of each arm. Finally, a joint position potential is used to keep the robot close to a safe configuration, similar to Fig. \ref{fig:setup}.

\subsection{Offline Goal-Conditioned  Reinforcement Learning}
\label{subsec:rl}
The Goal-Conditioned RL problem is formulated as a Markov Decision Process (MDP). We frame the task described in Section \ref{sec:problem_statement} as an episodic MDP, defined as a tuple $(\mathcal{S},\mathcal{A},\mathcal{G},p,r, \gamma)$, where $\gamma$ is the discount factor and $\mathcal{S}$, $\mathcal{A}$ and $\mathcal{G}$ are continuous state, action and goal spaces.  The probability density function $p(s_{t+1} | s_t, a_t)$ represents the probability of transitioning to state $s_{t+1}$, given the current state $s_t$ and action $a_t$, with $s_t, s_{t+1} \in \mathcal{S}$ and $a_t \in \mathcal{A}$. The dynamics of the interaction between the robot and the DLO $p(s_{t+1} | s_t, a_t)$ are unknown. Instead, real data is collected with the experimental setup shown in Fig. \ref{fig:setup}, which is used to learn a deterministic policy $\pi(s,g)=a$ i.e. the actor, based on a reward function $r: \mathcal{S} \times \mathcal{G} \rightarrow \mathbb{R}$, where $g$ indicates the desired goal. The return is defined as the sum of discounted future rewards: $G_t = \sum_{k=t}^{T} \gamma^{k-t} r (s_k,g_k)$, where $t$ and $T$ are the current and terminal state's indices, respectively. RL algorithms aim to maximize the expected return conditioned on state-action pairs, encoded as the action-value $Q(s,a,g)$ i.e. the critic, which also depends on the goal. 

In offline RL, the objective is for the agent to learn a policy based solely on a fixed dataset, $\mathcal{D}$. This is advantageous in robotics, however it also adds new challenges, given that agents tend to estimate the value of unseen state-action pairs incorrectly \cite{levine2020offline}. Offline GCRL further brings more difficulties, since there needs to be sufficient coverage of the goal space to have proper generalization \cite{yang2023essential}. In this work we employ the TD3+BC algorithm proposed by Fujimoto et al. \cite{fujimoto2021minimalist}, which aims to improve the value estimation problem by modifying the policy update step of the Twin Delayed DDPG (TD3) algorithm \cite{fujimoto2018addressing} with a Behavior Cloning (BC) regularization term: 
\begin{equation}
    \pi = \text{arg}\max_\pi \mathbb{E} \bigg[\lambda Q(s, \pi(s,g), g)  -\underbrace{(\pi(s,g) - a)^2}_{\text{BC regularization}}\bigg]
\end{equation}
where the expected value is computed over samples from the dataset, ${(s,a,g)\sim\mathcal{D}}$, and $\lambda$ is a weighting factor that regulates how much the BC term impacts the policy update, computed over batches of $P$ state-action-goal samples:
\begin{equation}
    \lambda= \frac{1}{P}\sum_{(s_i,a_i,g_i)}\frac{\alpha }{|Q(s_i, a_i ,g_i)|}
\end{equation}
where $\alpha$ is a hyperparameter which controls the strength of the regularization (larger $\alpha$ implies smaller impact of BC).

\subsubsection{\textbf{Data Collection}}
\begin{figure}[t]
    \centering
    \vspace{0.2cm}
    \includegraphics[width=\linewidth]{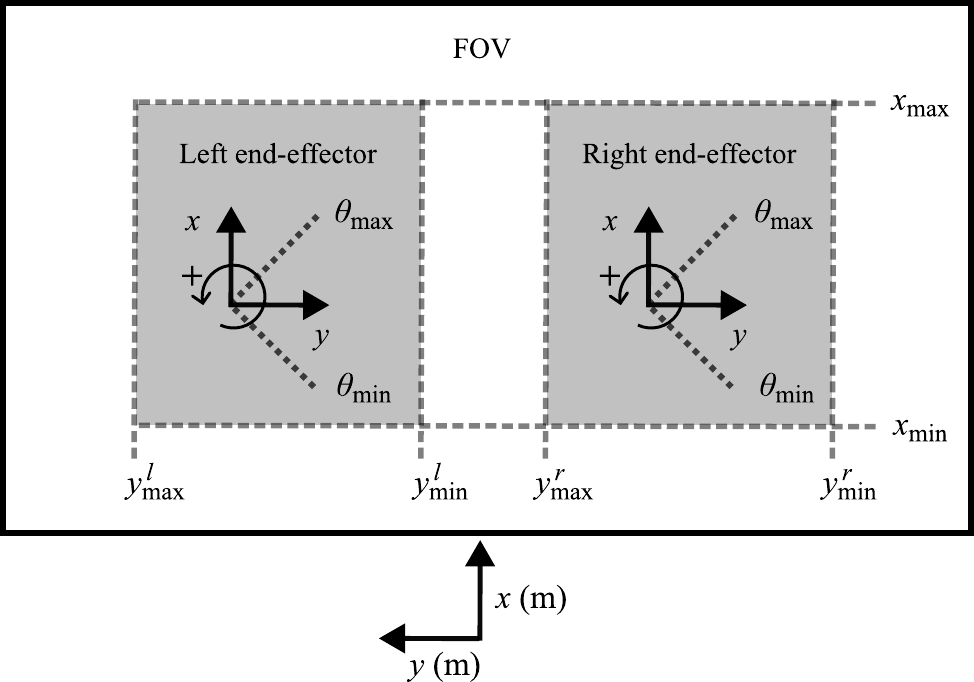} 
    \caption{Illustration of the safe workspace $\mathcal{W}_k$ in gray. The workspace of each end-effector is smaller than the field of view of the camera to prevent the DLO from leaving the visible region. Further, each end-effector stays in a separate region to prevent the DLO from becoming entangled. This also helps prevent end-effector collisions, making it unnecessary to include additional inequality constraints in the HQP formulation.}
    \label{fig:workspace}
    \vspace{-0.2cm}
\end{figure}

Since offline RL lacks the means for exploration, adequate coverage of the state-action space must be ensured to make learning feasible \cite{levine2020offline}. To that end, we developed a data collection procedure which deforms the rope into varied shapes. It works by randomly sampling positions from the safe workspace, $\mathcal{W}_k$ defined by ranges $[x_{\text{min}},x_{\text{max}}]$ and $[y^k_{\text{min}},y^k_{\text{max}}]$ where $k\in\{l,r\}$ indicates the left/right end-effector, and orientations from a range $[\theta_{\text{min}}, \theta_{\text{max}}]$. The safe workspace $\mathcal{W}_k$ is illustrated in Fig. \ref{fig:workspace} and consists of two mutually exclusive areas to prevent collisions and DLO entanglements. A simple point-to-point motion is then generated with a timing law dependent on the distance between the current and desired end-effector positions. Constraints are also added to keep the DLO inside the field of view (FOV) of the camera and prevent it from being overstretched. For each episode, there is a $0.3$ probability of sampling a new pose for the left, right or both end-effectors. With the remaining $0.1$ probability, a predetermined semi-random motion sequence is executed leading to an inversion of the DLO's curvature. Otherwise, the likelihood of observing this event completely at random is too low. Once each motion completes, the episode terminates and the final shape of the DLO is stored as the desired goal.

\subsubsection{\textbf{Data Augmentation}}
Besides covering the state-action space sufficiently, in GCRL it is equally important to ensure adequate coverage of the goal space \cite{yang2023essential}. To that end, we propose a simple data augmentation procedure that improves the performance of the offline RL method. The concept is similar to HER from online RL, which learns from failed experience \cite{andrychowicz2017hindsight}. In HER, after an episode is collected towards a goal $g$, additional goals $g'$ are sampled and used to recompute rewards, thus creating additional data to be stored in the replay buffer. In offline GCRL, the goal-conditioned policy, makes it possible to artificially generate new episodes by replacing the goal and recomputing the reward accordingly. This is particularly important to reduce the volume of experimental data needed. We further propose the following goal sampling procedures:
\begin{itemize}
    \item \textit{intra} - goals are sampled from intermediate shapes in the same episode, thus creating shorter episodes with a single point-to-point motion 
    \item \textit{inter} - goals are sampled from future episodes, thus creating longer episodes with multiple point-to-point motions 
    \item \textit{mixed} - goals are sampled from intermediate shapes in future episodes, thus creating longer episodes with multiple point-to-point motions 
\end{itemize}



\section{EXPERIMENTS}
\label{sec:experiments} 
The experimental setup is presented in Section \ref{subsec:exp_setup}, followed by three main experiments, described in Sections \ref{subsec:aug}, \ref{subsection:alpha} and \ref{subsection:material}, along with key results. Finally, limitations of the proposed method are summarized in Section \ref{subsection:limitation}.

\subsection{Experimental Setup} \label{subsec:exp_setup}
As shown in Fig. \ref{fig:setup}, we use an ABB dual-arm YuMi robot mounted with an Intel RealSense D435 camera fixed on top of the setup, facing the manipulation area. The test DLOs are fixed to the end-effectors of the YuMi using custom made 3D printed fingers. Two DLOs were tested, both with $1$ cm diameter and $55$ cm length: 
{\renewcommand{\theenumi}{(\textbf{\alph{enumi}}}
\begin{enumerate}
    \item a \textit{soft} rope [red]
    \item an \textit{elastic} cord [yellow]
\end{enumerate}}
The tracking procedure described in Section \ref{subsec:DLO_Tracking} was employed, using the respective DLO colors to produce the segmented images. In both cases, the lattice was initialized with $M=30$ cross-sections and the tracking output was set to have $N=18$ points. While the tracking procedure effectively works in three-dimensional space, for our purpose the $z$ dimension was ignored. This leads to a DLO state defined as $\pmb{q}^i\in\mathbb{R}^{18\!\times\!2}$, where $i\in\{c,d\}$ indicates the current and desired shapes. The position and orientation of the end-effectors is denoted by $\pmb{p}^i_j , \pmb{o}^i_j \in \mathcal{W}_k$, where $i\in\{c,d\}$ indicates the current and desired poses. Finally, $\mathcal{W}_k$ is defined with the following bounds: $[x_{\text{min}},x_{\text{max}}]=[0.3,0.6]$ m, $[y^l_{\text{min}},y^l_{\text{max}}] =[0.1,0.3]$ m, $[y^r_{\text{min}},y^r_{\text{max}}] = [-0.3,-0.1]$ m, and $[\theta_{\text{min}}, \theta_{\text{max}}]=\left[ -\frac{\pi}{4} ,\frac{\pi}{4} \right]$ rad. 

\subsubsection{\textbf{MDP Formulation}}
In practice, goal-conditioned formalism typically leads to an augmented actor and critic input vector, which consists of the state and goal concatenated together. In our implementation, the goal is prepended to the state, leading to the following input vector:
\begin{equation}
    g || s = [\bar{\pmb{q}}^d || \ \bar{\pmb{q}}^c, \  \pmb{p}^c_l, \  \pmb{p}^c_r, \  \pmb{o}^c_l, \  \pmb{o}^c_r] \in \mathbb{R}^{78}
\end{equation}
where $||$ denotes concatenation and the bar over the DLO shapes indicates flattened vectors, i.e. $\bar{\pmb{q}}^i\in \mathbb{R}^{36}$. The action space $\mathcal{A}$, is defined as the gripper poses, i.e. $a=[\pmb{p}^d_l, \  \pmb{p}^d_r, \  \pmb{o}^d_l, \  \pmb{o}^d_r]\in \mathbb{R}^6$. Finally, the reward function is defined based on the root mean squared error (RMSE):
\begin{equation}
    r(s|g) = - \sqrt{ \frac{1}{N} \textstyle \sum_{j=1}^{N} \left(\pmb{q}_j^d -\pmb{q}_j^c\right)^2} 
    \label{reward}
\end{equation}
The reward was chosen to be comparable with shape-servoing approaches which attempt to minimize the RMSE. 

\subsubsection{\textbf{Datasets}}
A small dataset of 1010 episodes was created for each DLO, with the last 10 episodes being used for testing alone and thus excluded from the training set. Data collection took approximately 3 hours for each dataset (episodes take $10.3\pm3.4$ s). Data is recorded at 20 Hz, but later down-sampled to 10 Hz to generate the datasets, as samples taken within 0.1 s show little changes in shape. For adequate comparison, the same point-to-point motions were used for both DLOs,  shown below:
\begin{figure}[h!]
    \centering
    \includegraphics[width=\linewidth]{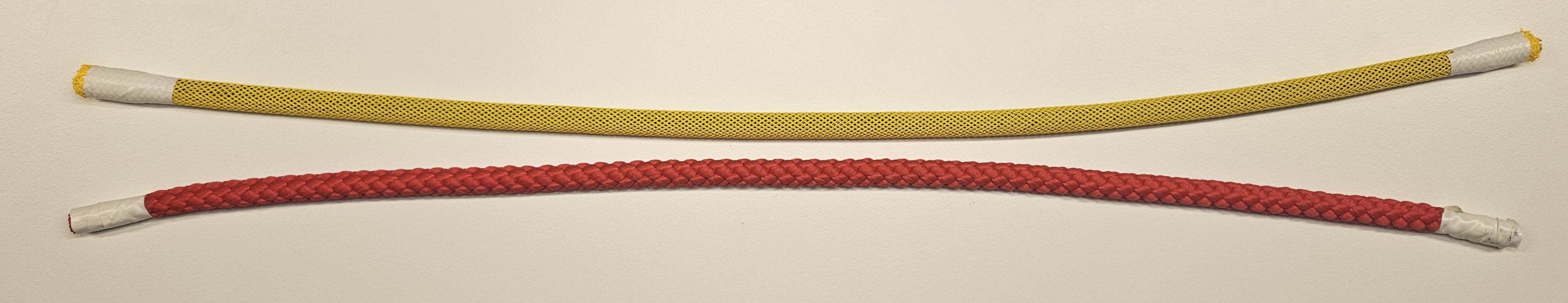} 
    \caption{Picture of the two types of DLO used to generate the datasets: a \textit{soft} rope (red) and an \textit{elastic} cord (yellow). Notably the elastic DLO was wrapped in a yellow plastic sleeve to enable better color segmentation and prevent longitudinal stretching.}
    \label{fig:dlos}
    \vspace{-0.1cm}
\end{figure}

The augmentation procedure described in Section \ref{subsec:aug} was performed on each dataset to generate additional goals.  Augmentation ratios are indicated by $\{\text{1x}, \text{2x}, \text{4x}, \text{6x}, \text{8x}\}$ where 1x refers to no augmentation, 2x indicates that the amount of data was doubled, i.e 2000 episodes, and so on.

\subsubsection{\textbf{Training Procedure}}
\looseness=-1
Policies are modeled as feed-forward ANNs using the D2RL architecture proposed by Sinha et al. \cite{sinha2020d2rl} which was shown to improve performance in robotic manipulation tasks. In this architecture, the state or the state-action pair are concatenated to the inputs of each hidden layer of the actor and critic networks, respectively. We use 4 hidden layers with 256 neurons each, and ReLU activation functions, except in the output layers which have linear activations. The Adam optimization algorithm was used, with default parameters ($\beta_1=0.9, \beta_2=0.999, \epsilon=1\times 10^{-8}$) and a learning rate of $0.0001$ for both actor and critic networks. Batch normalization and a $0.5$ dropout rate were used to stabilize the learning. Offline RL policies were trained using the TD3+BC algorithm, implemented in the \textit{d3rlpy} library \cite{seno2021d3rlpy}, with $1\times10^6$ environment steps and a 256 batch size. Note that earlier policies may be better, but finding a criteria for early stopping in offline RL is still an open research question, since there is no clear measure for over-fitting \cite{kumar2021workflow}. Lastly, a hyperparameter search resulted in $\gamma=0.95$.

\subsubsection{\textbf{Testing Procedure}}
To evaluate each policy, the system architecture in Fig. \ref{fig:schematics} is used. The current state is given as input to the ANN and its output is used to update the desired poses of the end-effectors, at 2 Hz. The controller operating at 50 Hz is constantly driving the end-effectors to the current desired poses. Each test starts with the DLO stretched along the $x=0.5$ m line (which is where the tracking is initialized). Then, a sequence of 8 test shapes determines the value of $\pmb{q}^d$, for a total of $40$ s (without a reset). Note that from the initial 10 test shapes, only the first 8 are reported here to provide a balanced test set with two straight shapes and three shapes of each curvature, i.e. convex and concave. For each DLO, the shape-servoing method proposed by Berenson \cite{berenson2013manipulation} was used as a comparative baseline. Note that the same testing methodology was used to test the baseline, and there was a non-exhaustive attempt to tune its parameters (e.g. diminishing rigidity value, $\mathtt{k}=1$).

\subsection{Data Augmentation}\label{subsec:aug}
In preliminary experiments with the \textit{soft} DLO, the proposed types of goal sampling resulted in the following average RMSEs and standard deviations (across 8 shapes):
\begin{itemize}
    \item \textit{intra} - RMSE$\ = 0.0381\pm0.0253$ m
    \item \textit{inter} - RMSE$\ = 0.0494\pm0.0314$ m
    \item \textit{mixed} - RMSE$\ = 0.0466\pm0.0318$ m
\end{itemize}
Notably, \textit{mixed} and \textit{intra} led to better results than the \textit{inter} sampling strategy, as they both add new goals to the dataset. The \textit{intra} strategy was used for the rest of the experiments as it led to a slightly lower RMSE, which is aligned with the findings by Andrychowicz et al. \cite{andrychowicz2017hindsight}. When comparing their \texttt{future}, \texttt{episode} and \texttt{random} goal sampling approaches, the conclusion was that the most valuable goals are those which are achieved in the near future. Indeed, the \textit{intra} augmentation procedure produced more episodes with shorter lengths, i.e. goals close to the starting state.

\begin{figure}[!t]
    \centering
    \includegraphics[width=\linewidth]{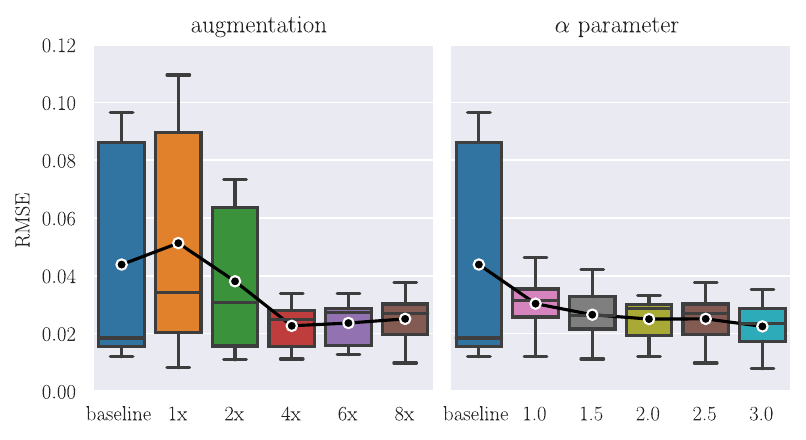} 
    \caption{Boxplots with results for the augmentation experiment (left) and the impact of BC regularization (right), determined by the $\alpha$ parameter. Note that the {\normalfont 8x} entry on the left is the same as the $2.5$ entry on the right, and so are the results for Berenson's baseline. The black dots indicate the mean across the 8 test shapes.}
    \label{fig:results}
    \vspace{-0.2cm}
\end{figure}

To evaluate the benefit of the augmentation procedure, four levels of augmentation were tested on the \textit{soft} DLO. The regularization parameter was fixed to $\alpha=2.5$, as in the original paper. From Fig. \ref{fig:results} (left), it is clear that data augmentation indeed helps achieve a lower average RMSE. This is likely due to the increase of goal shapes found in the augmented dataset. Notably, the positive effect seems to plateau after 4x augmentation, with an average
RMSE$\ =\{0.0227 \pm 0.0086, 0.0236 \pm 0.0082, 0.0251 \pm 0.0102\}$ m, for the ratios \{4x, 6x, 8x\}, respectively.

\begin{figure*}[t]
    \centering
    \includegraphics[width=\linewidth]{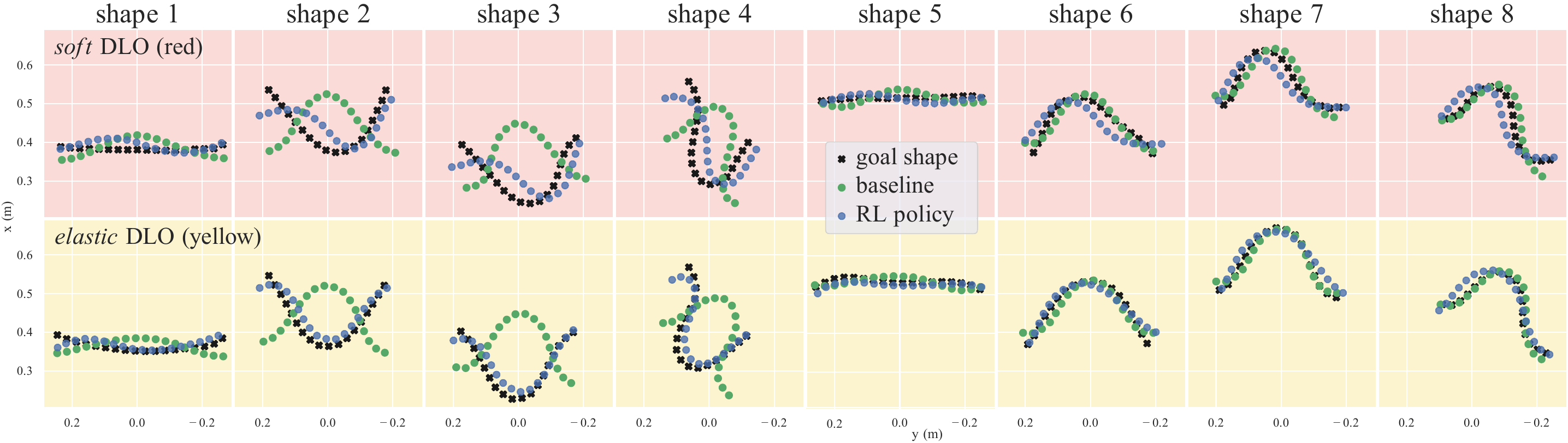} 
    \caption{Comparison between final test shapes and goal shapes. Note how the offline RL policy succeeded in inverting the curvature, while the shape-servoing baseline remained at a local minimum, for both DLOs: \textbf{(a)} soft DLO and  \textbf{(b)} elastic DLO. It is also possible to observe the differences in shapes due to the material properties, since the commanded end-effector poses are identical for both objects. In particular for the straight shapes, the middle part of the elastic DLO follows the end-effector angles more closely, when compared to the soft DLO. The test shapes with minimum and maximum RMSEs were: \textbf{(a)} baseline [shape 5, shape 4], RL policy [shape 5, shape 3]; \textbf{(b)} baseline [shape 7, shape 4], RL policy [shape 5, shape 4].}
    \label{fig:shapes}
    \vspace{-0.2cm}
\end{figure*}

\subsection{BC Regularization}\label{subsection:alpha}
Other offline RL algorithms were initially explored, but TD3+BC outperformed them all, including plain BC. While further hyperparameter tuning could change this outcome, the simplicity of TD3+BC was also appealing, with only one additional parameter to tune. In order to investigate the impact of the $\alpha$ regularization term, we used the 8x augmented dataset and varied the value of $\alpha$ for the \textit{soft} DLO. The results, shown in Fig \ref{fig:results} (right), indicate that larger values of $\alpha$ (i.e. decreasing impact of BC) lead to better results. The best result was achieved with $\alpha=3.0$, for an average 
RMSE$\ =0.0225\pm0.0093$ m, outperforming the baseline which in turn had an average 
RMSE$\ =0.0439\pm0.0394$ m.
This error difference is mostly due to the successful inversion of the DLO curvature for the initial shapes, shown in Fig. \ref{fig:shapes}.

\subsection{Material Properties}\label{subsection:material}
In order to evaluate the generalization ability of the proposed method, the best performing combination of augmentation ratio and BC regularization was applied to the \textit{elastic} DLO dataset, namely $\text{8x}$ and $\alpha=3.0$. The results for both objects are summarized in Table \ref{tab:results}. Notably, the baseline showed no significant difference across DLOs, while the RL approach performed better on the elastic DLO. This may be due to the simpler object dynamics of a stiffer material, as discussed in Section \ref{sec:problem_statement}. Fig. \ref{fig:shapes} also shows the effect of different material properties on the shape of the DLOs, as the same exact trajectories lead to different goal shapes.
\bgroup
\def\arraystretch{1.3}
\setlength\tabcolsep{3pt}
\begin{table}[h!]
\centering
\caption{Comparison of proposed method with Berenson baseline for two types of DLO material properties: \textbf{(a)} soft rope and \textbf{(b)} \textit{elastic} cord. Mean, standard deviation and ranges of RMSE values across the 8 test shapes (m).}
\begin{tabular}{c|cc|cc}
\thickhline
\multirow{2}{*}{\textbf{DLO}} & \multicolumn{2}{c|}{\textbf{Baseline}} & \multicolumn{2}{c}{\textbf{RL Policy}} \\
                  & mean $\pm \ \sigma$ & [min,max] & mean $\pm \ \sigma$ & [min,max] \\ \hline
\textbf{(a)} & $0.044\pm0.039$ & $[0.012,0.097]$ & $0.023\pm0.009$ & $[0.008,0.035]$ \\
\textbf{(b)} & $0.044\pm0.045$ & $[0.007,0.103]$ & $0.015\pm0.004$ & $[0.010,0.023]$ \\
\thickhline
\end{tabular}
\label{tab:results}
\end{table}
\bgroup


\subsection{Limitations}\label{subsection:limitation}
Although the proposed method has shown some positive results for the test sequence in Fig. \ref{fig:shapes}, it still struggles with consecutive shapes which are too different. Even reordering the test sequence leads to a decrease in performance. While this is also observed with the shape-servoing baseline which can only handle monotonically decreasing errors, RL should have the potential to learn more complex motions which do not satisfy this condition \cite{laezza2021learning}. While each RL policy was trained and tested once, it is a well known fact that there may be a large variability in the outcomes of separate learning trials \cite{henderson2018deep}. Therefore, more extensive tests need to be carried out to fully validate the results.

\section{CONCLUSIONS}
\looseness=-1
In this work, we investigate offline GCRL for shape control of DLOs with an ABB dual-arm YuMi robot, showcasing its potential to go beyond the capabilities of classical shape-servoing methods. In particular, we propose an approach using the TD3+BC algorithm alongside a data augmentation procedure inspired by the HER principle. Our experiments show the potential of using real-world data to allow effective manipulation of materials with different properties, such as soft ropes and elastic cords. Notably, the proposed method was able to handle curvature inversion of the DLO, in a simplified test sequence. 

\looseness=-1
Several important aspects remain to be explored, such as better state and action representations, the impact of the low-level control, the frequencies of both the tracking and control policies. As a real-robot learning problem, there are a lot of design choices related to the robotic system. Offline RL brings even more design freedom when it comes to algorithmic choices, ranging from macro decisions such as which RL method is used, to the micro decisions of learning hyperparameters, which all impact the learning outcome. 





\section*{ACKNOWLEDGMENTS}
This work was partially supported by the Wallenberg AI, Autonomous Systems and Software Program (WASP) funded by the Knut and Alice Wallenberg Foundation.
This work was also a part of project SOFTMANBOT, which received funding from the European Union's Horizon 2020 research and innovation program under grant agreement No 869855. Finally, the computations were enabled by resources provided by the National Academic Infrastructure for Supercomputing in Sweden (NAISS), partially funded by the Swedish Research Council through grant agreement no. 2022-06725.


\bibliographystyle{ieeetr}
\bibliography{references.bib}

\end{document}